\documentclass[runningheads]{llncs}

\usepackage{eccv}

\usepackage{eccvabbrv}

\usepackage{graphicx}
\usepackage{booktabs}

\usepackage[accsupp]{axessibility}  

\usepackage[pagebackref,breaklinks,colorlinks,citecolor=eccvblue]{hyperref}

\usepackage{orcidlink}

\begin{document}

\title{Improving the Multi-label Atomic Activity Recognition by Robust Visual Feature and Advanced Attention @ ROAD++ Atomic Activity Recognition 2024} 

\titlerunning{Abbreviated paper title}

\author{Jiamin Cao \and Lingqi Wang \and Kexin Zhang \and Yuting Yang \and \\
Licheng Jiao \and Yuwei Guo}

\authorrunning{F.~Author et al.}

\institute{Intelligent Perception and Image Understanding Lab, China
\email{23171214660@stu.xidian.edu.cn}
}
\maketitle

\begin{abstract}
Road++ Track3 proposes a multi-label atomic activity recognition task in traffic scenarios, which can be standardized as a 64-class multi-label video action recognition task. In the multi-label atomic activity recognition task, the robustness of visual feature extraction remains a key challenge, which directly affects the model performance and generalization ability. To cope with these issues, our team optimized three aspects: data processing, model and post-processing. Firstly, the appropriate resolution and video sampling strategy are selected, and a fixed sampling strategy is set on the validation and test sets. Secondly, in terms of model training, the team selects a variety of visual backbone networks for feature extraction, and then introduces the action-slot model, which is trained on the training and validation sets, and reasoned on the test set. Finally, for post-processing, the team combined the strengths and weaknesses of different models for weighted fusion, and the final mAP on the test set was 58\%, which is 4\% higher than the challenge baseline.  
\end{abstract}

\section{Introduction}
\label{sec:intro}

Atomic activity recognition in traffic scenarios is a crucial task in understanding and analyzing complex interactions in dynamic environments. The goal is to detect and classify multiple fine-grained actions that occur simultaneously or in quick succession. These actions, referred to as "atomic activities," can range from vehicle turning, lane changing, to pedestrian crossing, or any nuanced behaviors of road agents. The complexity of this task arises from the diversity and overlap of actions in a single video sequence, making it essential to develop robust models that can handle multiple labels for each frame.

Existing methods for atomic activity recognition can be broadly categorized into single-label and multi-label approaches. In single-label methods, the model assigns one dominant action to each frame or video segment, while multi-label approaches allow for the classification of multiple concurrent actions. Traditional approaches often rely on Convolutional Neural Networks (CNNs) and Recurrent Neural Networks (RNNs) to capture spatial and temporal dependencies, respectively. CNNs are used for spatial feature extraction from frames, while RNNs, particularly Long Short-Term Memory (LSTM) networks, handle temporal dependencies across frames. More recently, transformer-based architectures have gained traction due to their ability to capture long-range dependencies and handle multi-label classification tasks more effectively. Inspired by Francesco et. al \cite{locatello2020objectcentriclearningslotattention}, Kung et. al \cite{kung2024action}proposes a novel action-centered slot-attention-based framework that can decompose multiple atomic activities in a video.

However, atomic activity recognition presents additional challenges. In real-world traffic scenarios, the quality of visual feature extraction significantly impacts model performance and generalization. Poorly extracted features can result in degraded action classification, especially when multiple, overlapping activities are present. To address this, we optimize three aspects of the recognition pipeline, including data processing, model design, and post-processing techniques:
\begin{enumerate}
    \item Data Processing: The team emphasizes the importance of resolution and video sampling strategy. The correct resolution enhances feature extraction quality, while a fixed sampling strategy ensures consistency across validation and test sets. This step is crucial for handling temporal variations and ensuring that the model sees representative data from the entire video.
    \item Model Optimization: A variety of visual backbone networks are selected to maximize the robustness of feature extraction. By employing diverse backbones, the model can better generalize to complex traffic scenarios. Additionally, the introduction of an action slot model allows for targeted training on the training and validation sets, ensuring more effective inference on the test set. This model is designed to accommodate the nuances of multi-label classification, where multiple activities may overlap temporally.
    \item Post-Processing: After obtaining predictions from multiple models, the team performs weighted fusion to combine their strengths. This method allows for improved results by leveraging the complementary features of different models. The result is a significant boost in performance, achieving a mean Average Precision (mAP) of 58\% on the test set, surpassing the challenge baseline by 4\%.
\end{enumerate}

\section{Our Solution}

This track focus on the problem of atomic activity recognition within video clips $V_i =\{I^t_i\}^{T}_{t=1}$. Given a video clip consisting of $T$ consecutive image frames, our objective is to identify whether a set of predefined atomic activities $Y$, are present in the video. Atomic activities refer to the most fundamental action units observable in traffic scenes, which serve as the building blocks for more complex interactive activities.

For each video clip, we define a corresponding atomic activity label vector , where denotes the total number of possible atomic activity categories. Within this label vector, each element $y_c$ corresponds to a specific atomic activity category $c$, and is a binary value, where if the atomic activity of category $c$ is observed in the video, $y_c=1$, and otherwise $y_c = 0$.

We will introduce our solution in this section, which consists of three parts, data preprocessing, model, and post-processing. we will introduce data preprocessing at \ref{data proprecess}, including the processing of training and test data, the model at \ref{model}, and post-processing strategies at \ref{post-processing}.

\subsection{data processing}
\label{data proprecess}

To optimize the data processing pipeline, we carefully selected a strategy that balances efficiency with maintaining essential features for accurate recognition. First, the resolution of the dataset was reduced from 512x1536 to 256x658, significantly lowering the computational load without sacrificing critical visual details necessary for action recognition. Additionally, a fixed sampling strategy was employed for both the validation and test sets, ensuring consistency during evaluation. To further streamline the model input, the video sequences were downsampled to 16 frames, capturing sufficient temporal information while reducing redundancy, leading to more efficient training and inference.

\subsection{model}
\label{model}

In terms of model training, we proposed a robust visual feature extraction framework to capture inter-frame dependencies more effectively. For video-level feature extraction, we selected advanced backbone networks such as X3D\cite{feichtenhofer2020x3d} and SlowFast\cite{feichtenhofer2019slowfast}, which excel in learning both fine-grained motion dynamics and temporal relationships across frames. Additionally, we incorporated slot attention to focus on learning action-centric representations, enabling the model to capture not only the motion within the video but also the broader contextual information. This approach enhances the model's ability to recognize complex, overlapping activities in dynamic traffic environments.

\subsection{post-processing}
\label{post-processing}

For post-processing, we employed model ensemble techniques to integrate the outputs of different backbone models. Specifically, we performed a weighted sum of the similarity matrices generated by each model, combining their individual strengths to produce the final prediction. This approach leverages the complementary features learned by each model, enhancing the overall robustness and accuracy of the results.

\section{Experiments}
\subsection{Dataset}

TACO dataset, which consists of 13 video scene folders that include different maps, collected in CARLA simulator and different collection methods (Autopilot, Scene Runner, and manual collection of all three), where the size of each image frame is 512 x 1536. the dataset has been divided into training, validation, and test sets. The TACO dataset consists of a total of 5178 video clips out of which 1148 are used for testing. Each clip collects images and instance segmentation data at an initial resolution of 512 × 1536 pixels.
\subsection{Evaluation}
The evaluation metric used for this task is the commonly adopted mean Average Precision (mAP), which is frequently applied in multi-label video recognition tasks. In addition to overall mAP, we also report the agent-specific mAP scores, including four-wheeled vehicles (mAP@c), two-wheeled vehicles (mAP@k), pedestrians (mAP@p), grouped four-wheeled vehicles (mAP@c+), grouped two-wheeled vehicles (mAP@k+), and grouped pedestrians (mAP@p+). These metrics provide a more detailed performance assessment across different types of road agents.
\subsection{Results}

We performed training on the TACO dataset by training the training set and the validation set using three backbones, x3d-l,x3d-m, and slow-fast, which were trained for 100epoch or 150epoch, respectively, and the rest of the settings were kept the same as action-slot. Experiments results in \ref{tab:result}.

\begin{table}
    \centering
    \begin{tabular}{cccccccc}
        \midrule
          & mAP & mAP@C &mAP@K  &mAP@P  &mAP@C+  &mAP@K+ &mAP@P+ \\
         \midrule
         baseline&0.54  &0.48  &0.41  &0.49  &0.70  &0.62 &0.53  \\
         x3d-l-100e-train&0.45  &0.42  &0.34  &0.42  &0.58  &0.50 &0.45 \\
         x3d-l-150e-trainval&0.49  &0.42  &0.35  &0.42  &0.66  &0.58 &0.52 \\
         slow-fast-150e-trainval&0.40  &0.37  &0.25  &0.41  &0.56  &0.43 &0.43 \\
         fusion& \textbf{0.58}  &\textbf{0.51}  &\textbf{0.47}  &\textbf{0.51}  &\textbf{0.74}  &\textbf{0.66} &\textbf{0.56} \\
         \midrule
    \end{tabular}
    \caption{Results on TACO test set}
    \label{tab:result}
\end{table}

\section{Conclusion}

For the multi-label atomic activity recognition task, the team optimized the improvement in three areas: data processing, model training and post-processing. We aim to perform robust visual feature extraction and advanced attention mechanisms as well as model ensemble, which proves the effectiveness of our optimization by improving the mAP by 4\% on the test set compared to the baseline results.

%
%
\bibliographystyle{splncs04}
\bibliography{main}

\begin{thebibliography}{1}
\providecommand{\url}[1]{#1}
\csname url@samestyle\endcsname
\providecommand{\newblock}{\relax}
\providecommand{\bibinfo}[2]{#2}
\providecommand{\BIBentrySTDinterwordspacing}{\spaceskip=0pt\relax}
\providecommand{\BIBentryALTinterwordstretchfactor}{4}
\providecommand{\BIBentryALTinterwordspacing}{\spaceskip=\fontdimen2\font plus
\BIBentryALTinterwordstretchfactor\fontdimen3\font minus \fontdimen4\font\relax}
\providecommand{\BIBforeignlanguage}[2]{{%
\expandafter\ifx\csname l@#1\endcsname\relax
\typeout{** WARNING: IEEEtran.bst: No hyphenation pattern has been}%
\typeout{** loaded for the language `#1'. Using the pattern for}%
\typeout{** the default language instead.}%
\else
\language=\csname l@#1\endcsname
\fi
#2}}
\providecommand{\BIBdecl}{\relax}
\BIBdecl

\bibitem{locatello2020objectcentriclearningslotattention}
\BIBentryALTinterwordspacing
F.~Locatello, D.~Weissenborn, T.~Unterthiner, A.~Mahendran, G.~Heigold, J.~Uszkoreit, A.~Dosovitskiy, and T.~Kipf, ``Object-centric learning with slot attention,'' 2020. [Online]. Available: \url{https://arxiv.org/abs/2006.15055}
\BIBentrySTDinterwordspacing

\bibitem{kung2024action}
C.-H. Kung, S.-W. Lu, Y.-H. Tsai, and Y.-T. Chen, ``Action-slot: Visual action-centric representations for multi-label atomic activity recognition in traffic scenes,'' in \emph{Proceedings of the IEEE/CVF Conference on Computer Vision and Pattern Recognition}, 2024, pp. 18\,451--18\,461.

\bibitem{feichtenhofer2020x3d}
C.~Feichtenhofer, ``X3d: Expanding architectures for efficient video recognition,'' in \emph{Proceedings of the IEEE/CVF conference on computer vision and pattern recognition}, 2020, pp. 203--213.

\bibitem{feichtenhofer2019slowfast}
C.~Feichtenhofer, H.~Fan, J.~Malik, and K.~He, ``Slowfast networks for video recognition,'' in \emph{Proceedings of the IEEE/CVF international conference on computer vision}, 2019, pp. 6202--6211.

\end{thebibliography}
\end{document}